\documentclass[10pt,twocolumn,letterpaper]{article}

\usepackage{cvpr}
\usepackage{times}
\usepackage{epsfig}
\usepackage{graphicx}
\usepackage{amsmath}
\usepackage{amssymb}

\usepackage{subfigure}
 \cvprfinalcopy % *** Uncomment this line for the final submission

\def\cvprPaperID{8252} % *** Enter the CVPR Paper ID here

\ifcvprfinal\fi
\begin{document}

%%%%%%%%% TITLE
\title{Two-Stage Peer-Regularized Feature Recombination \\for Arbitrary Image Style Transfer}
\author{
  Jan Svoboda\textsuperscript{1,2}, Asha Anoosheh\textsuperscript{1}, Christian Osendorfer\textsuperscript{1}, Jonathan Masci\textsuperscript{1} \\
  \textsuperscript{1}NNAISENSE, Switzerland \hspace{5mm} 
  \textsuperscript{2}Universita della Svizzera italiana, Switzerland \hspace{5mm} \\
  {\tt\small \{jan.svoboda,asha.anoosheh,christian.osendorfer,jonathan.masci\}@nnaisense.com} 
   }
% \and

\maketitle
%\thispagestyle{empty}

%%%%%%%%% ABSTRACT
\begin{abstract}
  This paper introduces a neural style transfer model to generate a stylized image conditioning on a set of examples describing the desired style.
   The proposed solution produces high-quality images even in the zero-shot setting and allows for more freedom in changes to the content geometry.
  This is made possible by introducing a novel Two-Stage Peer-Regularization Layer that recombines 
  style and content in latent space by means of a custom graph convolutional layer. 
  %aiming at separating style and content.
  Contrary to the vast majority of existing solutions, our model does not depend on any pre-trained networks for
  computing perceptual losses and can be trained fully end-to-end thanks to a new set of cyclic losses that
  operate directly in latent space and not on the RGB images. 

  An extensive ablation study confirms the usefulness of the proposed losses and of the 
  Two-Stage Peer-Regularization Layer, with qualitative results that are competitive with respect to
  the current state of the art using a single model for all presented styles. 
  %Moreover, we demonstrate that our approach is competitive even in the challenging zero-shot setting. 
  This opens the door to more abstract and artistic neural image generation
  scenarios, along with simpler deployment of the model.
\end{abstract}

%%%%%%%%% BODY TEXT
\section{Introduction}

Neural style transfer (NST), introduced by the seminal work of Gatys~\cite{Gatys2015NeuralStyle}, is an area of 
research that focuses on models that transform the visual appearance of an input image (or video) to match the style of a desired target image. 
For example, a user may want to convert a given photo to appear as if Van Gogh had painted the scene.

NST has seen a tremendous growth within the deep learning community and spans a wide spectrum of applications e.g.\ converting time-of-day~\cite{Zhu2017CycleGAN, Huang2018MUNIT}, mapping among artwork and photos~\cite{Anoosheh2017ComboGAN, Zhu2017CycleGAN, Huang2018MUNIT}, transferring facial expressions~\cite{Karras2018StyleBasedGen}, transforming animal species~\cite{Zhu2017CycleGAN, Huang2018MUNIT}, etc.
%Neural style transfer~(NST) is an appealing area of research allowing us to generate new images based on an existing image we have and some desired properties of the output we aim to generate, which can be specified by a set of one or more so called style images. As an example, one could take a photo and wish to convert it into a painting by one of his favorite painters, preserving the content of the photo he has taken, but making it look like as if it was painted by, say, Van Gogh. Some other applications can involve for example changing night photo into day, sketches into real photos, etc.

\begin{figure}[ht!]
\centering
\includegraphics[scale=0.19]{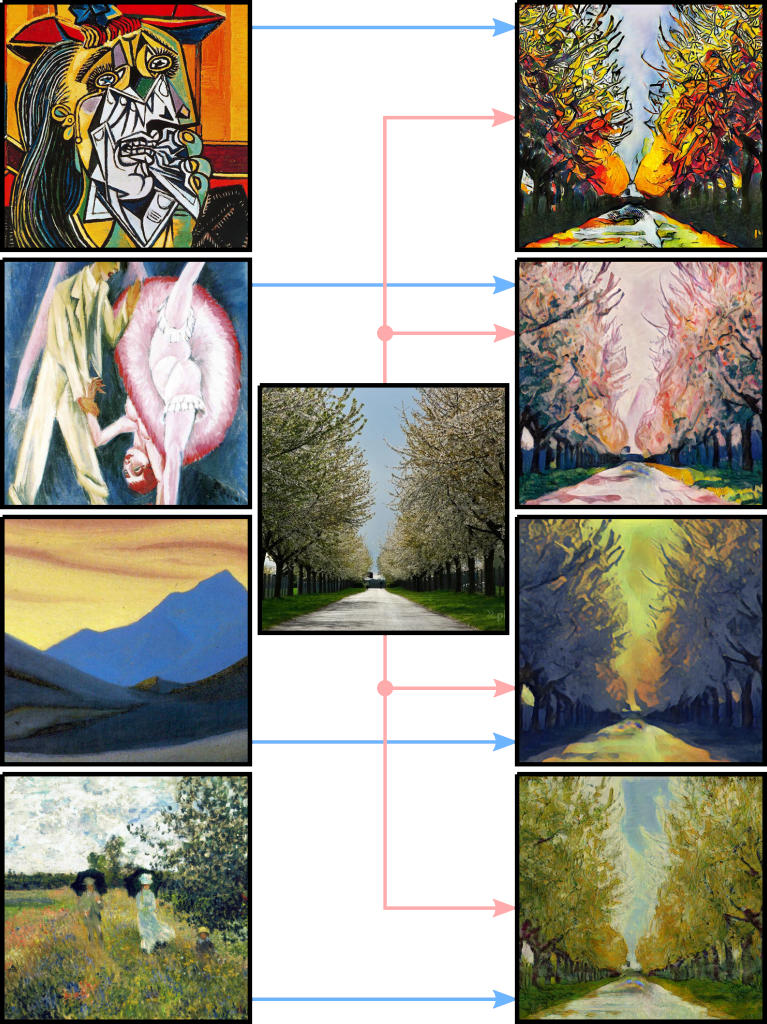}
\caption{Our model is able to take a content image and convert it into arbitrary target style in a forward manner. In this figure, a photo in the middle is converted into styles from 4 different painters, (left side, from top to bottom) Picasso, Kirchner, Roerich and Monet. }
\label{fig:teaser}
\vspace{-3mm}
\end{figure}
Despite their popularity and often-quality results, current NST approaches are not free from limitations. 
Firstly, the original formulation of Gatys \etal~requires a new optimization process for each transfer performed, 
making it impractical for many real-world scenarios. In addition, the method relies heavily on pre-trained 
networks, usually borrowed from classification tasks, that are known to be sub-optimal and have recently been
shown to be biased toward texture rather than structure~\cite{Geirhos2018ImageNetTrained}.
To overcome this first limitation, 
deep neural networks have been proposed to approximate the lengthy optimization procedure in a 
single feed forward step
% research has been done using deep neural networks to approximate the lengthy optimization procedure in a single feed forward step,
thereby making the models amenable for real-time processing. 
Of notable mention in this regard are the works of Johnson \etal~\cite{Johnson2016PerceptLosses} and Ulyanov \etal~\cite{Ulyanov2016}.
% , who also later introduced
% Instance Normalization~\cite{Ulyanov17IN}, a popular feature-normalization scheme for style transfer.
%}... [here needs some text]

Secondly, when a neural network is used to overcome the computational burden of~\cite{Gatys2015NeuralStyle}, 
training of a model for every desired style image is required due to the limited-capacity of conventional
models in encoding multiple styles into the weights of the network.
% , due to the fact that the style information is being encoded directly into limited-capacity model weights.
%[some text to justify why, cannot encode all possible transformations in the weights?].
This greatly narrows down the applicability of the method for use cases where the concept of style 
cannot be defined a-priori and needs to be inferred from examples.
% is fuzzy and is perceived as features across multiple images instead of just one.
With respect to this second limitation, recent works attempted to separate style and content 
in feature space (latent space) to allow generalization to a style characterized by an additional input image,
or set of images.
The most widespread work in this family is AdaIN~\cite{Huang2017AdaIN}, a specific case of
FiLM~\cite{Perez2017FiLM}. 
The current state of the art allows to control the amount of stylization applied, 
interpolating between different styles, and using masks to convert different regions of image into different 
styles~\cite{Huang2017AdaIN, Sheng2018Avatar}.%\todo{how does this last sentence fit with the rest? cites or remove}

%Neural style transfer was first introduced by the seminal work of Gatys~\cite{Gatys2015NeuralStyle} achieving very impressive results at the time. The method however has its limitations. It can produce decent results, but it has to run new optimization process for each single style transfer, which makes it very impractical. Nevertheless, it has proposed some ideas that are used throughout the research up till nowadays. Many works have followed trying to remove the computational burden of the first method. Some researchers proposed systems which are able to convert any input into particular style. Still, one model per style has to be trained. Other works concentrated on trying to separate the style and content information in the image, proposing systems which could take couple of images, one for style and one for content, and produce the desired result. The most widespread work from this family is the so called AdaIN~\cite{Huang2017AdaIN}. It is basically a particular case of FiLM~\cite{Perez2017FiLM} method introduced later the same year. The current state-of-the-art allows, besides other things, to control the amount of stylization applied, interpolate between different styles, use masks to convert different regions of image into different styles, etc.

Beyond the study of new network architectures for improving NST, research has resulted in better suited loss functions to train the models. 
The perceptual loss~\cite{Gatys2015NeuralStyle,Johnson2016PerceptLosses} with a pre-trained VGG19 
classifier~\cite{SimonyanZ14} is very commonly used in this task as it, supposedly, captures high-level
features of the image. 
However, this assumption has been recently challenged in~\cite{Geirhos2018ImageNetTrained}. 
% Instead, similar to Cycle-GAN~\cite{Zhu2017CycleGAN} our model is trained from scratch on the image 
% style transfer problem with the hope of obtaining features more suitable for the task at hand.
Cycle-GAN~\cite{Zhu2017CycleGAN} proposed a new cycle consistent loss that does not require one-to-one 
correspondence between the input and the target images, thus lifting the heavy burden of data annotation.

The problem of image style transfer is challenging, because the style of an image is expressed by both \textit{local} properties~(e.g. typical shapes of
objects, etc.) and \textit{global} properties~(e.g. textures, etc.). 
Of the many approaches for modelling the content and style of an image that have been proposed in the past, encoding of the information in a lower dimensional latent space has shown very promising results.
We therefore advocate to model this hierarchy in latent space by local aggregation of pixel-wise features and by
the use of metric learning to separate different styles.
% This can be addressed in the latent space by aggregating part of the pixel-wise latent space for each feature map. In order to separate different styles in the latent space, one could use, for example, a metric learning approach. 
To the best of our knowledge, this has not been addressed by previous approaches explicitly.%\todo{this paragraph is confusing and should be made clear. how do local and global properties fit in our framework?}

%[A bit mixed with background, here one would have to cite CycleGAN etc]
%[Here needs bridge to motivate our approach. Why is it needed? Encode style indirectly rather than directly in the weights, therefore avoiding the need of multiple encoders and decoders. Btw the fact that multiple nets are needed was never mentioned explicitly.]

In the presence of a well structured latent space where style and content are fully separated, 
transfer can be easily performed by exchanging the style information in latent space between the
input and the conditioning style images, without the need to store the transformation in the decoder weights.
% one can train a model that encodes the input image and reconstructs it from the latent code. 
% The transfer can then be performed transparently by style exchange in the latent space, conditioned by an input style image, without relying on stylization being learned in the decoder weights. 
Such an approach is independent with respect to feature normalization and further avoids the need for rather
problematic pre-trained models.

However, the content and style of an image are not completely separable. The content of an image exhibits changes in geometry depending on what style it is painted with. Recently, Kotovenko~\etal~\cite{Kotovenko2019ContentTransformer} has proposed a content transformer block in an adversarial setting, where the model is trained in two stages. First a style transfer network is optimized. Then it is fixed and content transformer block is optimized instead, learning to account for changes in geometry relevant to a given style. The style exchange therefore becomes two-stage. Modeling such dependence has been shown to improve the visual results dramatically. 

%\One of the significant limitations of the current approaches is the use of so called perceptual loss based typically on features from pre-trained VGG-19 image classification model~\cite{SimonyanZ14}. Such models have been lately shown to over-concentrate on the texture information~\cite{Geirhos2018ImageNetTrained} and it is therefore questionable whether their use for perceptual loss is correct. It could also be that the only reason it works is actually that the models overly focus on the textures. 
%Moreover, such approaches might result in style-transfer that is focusing on changing the texture of the content image accordingly, without necessarily producing a result that would have the desired style properties. Let's take, say, Picasso's paintings. Many current methods will produce results that will seemingly look Picasso-like, but will lack some basic properties of Picasso's painting style regarding brush strokes, geometry, etc.

This paper addresses the NST setting where style is externally defined by a set of input images to allow transfer 
from arbitrary domains and to tackle the challenging zero-shot style-transfer scenario by introducing a novel 
feature regularization layer capable of recombining global and local style content from the input style image. It is achieved by borrowing ideas from geometric deep learning~(GDL)~\cite{Bronstein2017Geometric} and modelling pixel-wise graph of peers on the feature maps in the latent space. 
%This inductive bias is shown to allow the network to learn how to separate style and content rather than encoding the transformation in its weights. 
To the best of our knowledge, this is the first work that successfully leverages the power of GDL in the style transfer scenario.
We successfully demonstrate this in a series of zero-shot style transfer experiments, 
whose generated result would not be possible if the style was not actually inferred from the respective input images.

This work addresses the aforementioned limitations of NST models by making the following contributions:
%[last paragraph needs rewrite but after bridge text will be easier i guess]

%Besides, there have been attempts to completely separate style and content information of an image or a painting. We might ask ourselves whether this is semantically correct. It is true that for many painters, the style they paint differs based on the content of their painting. They paint sky in one way, trees and buildings or people in another. It might be therefore incorrect to assume that style and content are two pieces of information we should treat in a fully disentangled way. A partial disentanglement, while keeping some of the local style information related to content could be an alternative.

%Compared to the state-of-the-art, we believe that we can bring novelty into image style transfer by considering what was mentioned in the above paragraphs. We propose a novel method based on PeerNets architecture~\cite{Svoboda2019}, which has shown how to build a graph structure on an intermediate feature embeddings. We train a GAN-like architecture, with the aim to partially separate the content and style information in the latent space. The style part of the latent code is then reconstructed using feature space of the style image by means of modified Peer Regularization layer~\cite{Svoboda2019}. Our main contributions are:
\begin{itemize}
    \item A state-of-the-art approach for NST using a custom graph convolutional layer that recombines 
  style and content in latent space;
    \item Novel combination of existing losses that allows end-to-end training without the need for any pre-trained model~(e.g. VGG) to compute perceptual loss;
    \item Constructing a globally- and locally-combined latent space for content and style information and imposing structure
    on it by means of metric learning.
    %\item Computing transfer loss on latent representations, allowing for more freedom in the resulting geometry of the stylized image.\todo{this last point should somehow be merged with second}
    %\item no direct loss on the stylized RGB image, yielding more freedom in its structure. 
\end{itemize}

\section{Background}
The key component of any NST system is the modeling and extraction of the "style" from an image (though the term
is partially subjective). As style is often related to texture, a natural way to model it is to use
visual texture modeling methods~\cite{PaulyGreiner2009TexSynth}. Such methods can either exploit texture image 
statistics~(e.g. Gram matrix)~\cite{Gatys2015NeuralStyle} or model textures using Markov Random 
Fields~(MRFs)~\cite{EfrosLeung1999}. 
The following paragraphs provide an overview of the style transfer 
literature introduced by~\cite{Jing2017NSTReview}.%\todo{did you rephrase it or took it as is?}.

%Such methods are typically divided into two groups~\cite{Jing2017NSTReview}. The \textit{Parametric Texture Modelling with Summary Statistics} captures the images statistics from a sample texture and exploits summary statistical property to model the texture, filter responses or e.g. Gram-matrix based representations~\cite{Gatys2015NeuralStyle}. Second group is \textit{Non-parametric Texture Modelling with MRFs} which does, for example, synthesis of each pixel one-by-one by searching similar neighborhoods in the source texture and assigning the corresponding pixel~\cite{EfrosLeung1999}.

%Knowing how to treat the target style information, one can use image reconstruction techniques to reconstruct back image with the original content and desired style. Image reconstruction methods can be divided into \textit{Image-Optimisation-Based Online Image Reconstruction}, which is based on gradient descent in the image space, and \textit{Model-Optimisation-Based Offline Image Reconstruction}, which addresses some efficiency issues of the former.

\vspace{-1mm}
\paragraph{Image-Optimization-Based Online Neural Methods.} 
The method from Gatys \etal~\cite{Gatys2015NeuralStyle} may be the most representative of this category. While experimenting with representations from intermediate layers of the VGG-19 network, the authors observed that a deep convolutional network is able to extract image content from an arbitrary photograph, as well as some appearance information from works of art. The content is represented by a low-level layer of VGG-19, whereas the style is expressed by a combination of activations from several higher layers, whose statistics are described using the network features’ Gram matrix. 
Li~\cite{Li2017DemystifyingNST} later pointed out that the Gram matrix representation can be generalized using a formulation based on Maximum Mean Discrepacy~(MMD). Using MMD with a quadratic polynomial kernel gives results very similar to the Gram matrix-based approach, while being computationally more efficient. Other non-parametric approaches based on MRFs operating on patches were introduced by Li and Wand~\cite{LiWand2016MRF}.
\vspace{-5mm}
\paragraph{Model-Optimization-Based Offline Neural Methods.} These techniques can generally be divided into several sub-groups~\cite{Jing2017NSTReview}. \textit{One-Style-Per-Model} methods need to train a separate model for each new target style~\cite{Johnson2016PerceptLosses, Ulyanov2016, LiWand2016Adversarial}, rendering them rather impractical for dynamic and frequent use. A notable member of this family is the work by Ulyanov \etal~\cite{Ulyanov17IN} introducing Instance Normalization~(IN), better suited for style-transfer applications than Batch Normalization~(BN).
%Typical examples are works by Johnson~\cite{Johnson2016PerceptLosses} and Ulyanov~\cite{Ulyanov2016}, who later found out that it is more beneficial for neural style transfer to use so called Instance Normalization~(IN)~\cite{Ulyanov17IN} in place of more traditional Batch Normalization~(BN). A related, non-parametric patch-based approach using MRFs was proposed by Li and Wand~\cite{LiWand2016Adversarial}.

\textit{Multiple-Styles-Per-Model} methods attempt to assign a small number of parameters to each style. Dumoulin~\cite{Dumoulin2016CIN} proposed an extension of IN called Conditional Instance Normalization~(CIN), StyleBank~\cite{Chen2017StyleBank} learns filtering kernels for different styles, 
and other works instead feed the style and content as two separate inputs~\cite{Li2017, ZhangDana2017} 
similarly to our approach.

\textit{Arbitrary-Style-Per-Model} methods either treat the style information in a non-parametric, i.e. as in StyleSwap~\cite{Chen2016StyleSwap}, or parametric manner using summary statistics, such as in Adaptive Instance Normalization~(AdaIN)~\cite{Huang2017AdaIN}. 
AdaIN, instead of learning global normalization parameters during training, uses first moment statistics
of the style image features as normalization parameters.
% Instead of learning the normalization parameters during model training, AdaIN encodes the set of normalization parameters by using first moment statistics of the features for a given style image. 
Later, Li \etal~\cite{Li2017WCT} introduced a variant of AdaIN using Whitening and Coloring Transformations~(WTC).
Going towards zero-shot style transfer, ZM-Net~\cite{WangLiang2017ZMNet} proposed a transformation network with
dynamic instance normalization to tackle the zero-shot transfer problem. More recently, Avatar-Net~\cite{Sheng2018Avatar} proposed the use of a "style decorator" to re-create 
content features by semantically aligning input style features with those derived from the style image.%\todo{still correct after rewrite?}

\vspace{-1mm}
\paragraph{Other methods.} Cycle-GAN~\cite{Zhu2017CycleGAN} introduced a cycle-consistency loss on the reconstructed images that delivers very appealing results without the need for aligned input and target pairs. 
However, it still requires one model per style. 
The approach was extended in Combo-GAN~\cite{Anoosheh2017ComboGAN}, which lifted this limitation and allowed for a practical multi-style transfer; however, also this method requires a decoder-per-style.
% and results in a large
% model when training many different styles.

Sanakoyeu \etal~\cite{Sanakoyeu2018StyleAware} observed that the applying the cycle consistency loss in image
space might be over-restricting the stylization process.  They also show how to use a set of images, rather than a single one, to better express the style of an artwork. In order to provide more accurate style transfer with respect to an image content, Kotovenko \etal~\cite{Kotovenko2019ContentTransformer} designed the so called content transformer block, an additional sub-network that is supposed to finetune a trained style transfer model to a particular content. The strict consistency loss was later relaxed by 
MUNIT~\cite{Huang2018MUNIT}, a multi-modal extension of UNIT~\cite{Liu2017UNIT}, which imposes it in latent space
instead, providing more freedom to the image reconstruction process. Later, the same authors proposed FUNIT~\cite{Liu2019FUNIT}, a few-shot extension of MUNIT.

Towards exploring the disentanglement of different styles in the latent space, Kotovenko \etal~\cite{Kotovenko2019StyleDisentangle} proposed a fixpoint triplet loss in order to perform metric learning in the style latent space, showing how to separate two different styles within a single model.

\section{Method}
\label{sec:Method}
The core idea of our work is a region-based mechanism that exchanges the style between input and target style images,
similarly to StyleSwap~\cite{Chen2016StyleSwap}, while preserving the semantic content.
% Similarly to StyleSwap~\cite{Chen2016StyleSwap}, our idea is based upon on region-based style exchange between the content and the style image. 
To successfully achieve this, style and content information must be well separated, disentangled.
%The inductive bias of the network architecture is, however, not enough to achieve the desired level of separation. 
We advocate the use of metric learning to directly enforce separation among different styles, which has been experimentally shown to greatly reduce the amount of style dependent information retained in the decoder.
% Ideally, however, one should not only desire an exchange of style between source and target domains, but also the separation of style and content information in the latent space in order to transfer style without affecting the content significantly.
Furthermore, in order to account for geometric changes in content that are bound to a certain style, we model the style transfer as a two-stage process, first performing style transfer, and then in the second step modifying the content geometry accordingly. This is done using the Two-stage Peer-regularized Feature Recombination~(TPFR) module presented below.
\subsection{Architecture and losses}
\label{subsec:Arch}
\begin{figure*}[ht!]
\centering
\includegraphics[scale=0.375]{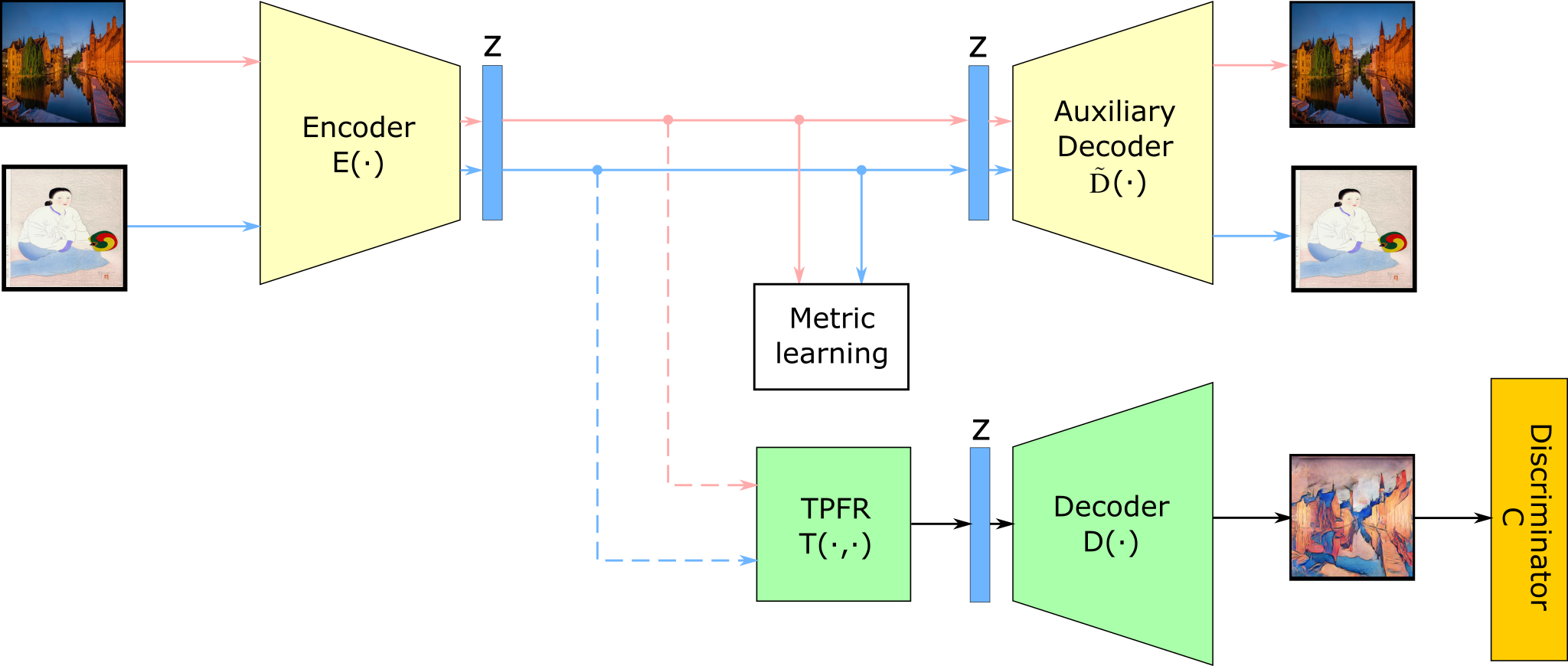}
\caption{The proposed architecture with two decoder modules. The yellow decoder is the auxiliary decoder, while the main decoder is depicted in green. Dashed lines indicate lack of gradient propagation.}
\label{fig:architecture}
\end{figure*}

The proposed system architecture is shown in Figure~\ref{fig:architecture}. To prevent the main decoder from encoding the stylization in its weights, auxiliary decoder~\cite{DeFauw2019AuxDec} is used during training to optimize the 
parameters of the encoder and decoder independently. The yellow module in Figure~\ref{fig:architecture} is trained as an 
autoencoder~(AE)~\cite{Masci2011,ZhaoMGL15,MaoSY16a} to reconstruct the input. 
The green module, instead, is trained as a GAN\cite{Goodfellow2014GAN} to generate the stylized version of the 
input using the encoder from the yellow module, with fixed parameters.
% At the same time, the green module is trained as GAN to produce stylized version of the input, borrowing the encoder from the yellow module with its parameters fixed. 
The optimization of both modules is interleaved together with the discriminator.
Additionally, following the analysis from Martineau \etal.~\cite{Martineau2018RAGAN}, the Relativistic Average GAN~(RaGAN) is used as our adversarial loss formulation, which has shown to be more stable and to produce more natural-looking images than traditionally used GAN losses.

Let us now describe the four main building blocks of our approach in detail\footnote{The full architecture details
are found in the supplementary material.}. 
We denote $x_i, x_t, x_f$ an input image, a target and fake image, respectively. 
Our model consists of an encoder $E(\cdot)$ generating the latent representations, an auxiliary decoder $\tilde{D}(\cdot)$ taking a single latent code as input, a main decoder $D(\cdot)$ taking one latent code as input, and TPFR module $T(\cdot, \cdot)$ receiving two latent codes. Generated latent codes are denoted $z_i = E(x_i), z_t = E(x_t)$. We further denote $(\cdot)_C, (\cdot)_S$ the content and style part of the latent code, respectively. 

The distance $f$ between two latent representations is defined as the smooth L1 norm~\cite{Huber1964}
in order to stabilize training and to avoid exploding gradients:
\begin{equation}
\begin{split}
    f[d] = \frac{1}{W \times H}\sum_{i=1}^{W \times H} \sum_{j=1}^{N} d_{i,j},\\
    d_{i,j} = \begin{cases} 0.5 \, d_{i,j}^2 & \text{if } d_{i,j} < 1 \\
                            |d_{i,j}| - 0.5 & \text{otherwise}      %
        \end{cases},
\end{split}
\end{equation}
where $d=z_1-z_2$, $z_1$ and $z_2$ are two different feature embeddings with $N$ channels and $W \times H$ is the spatial dimension of each channel.

%\todo{x used already, need notation for features}

\paragraph{Encoder.} The encoder used to produce latent representation of all input images is composed of several 
strided-convolutional layers for downsampling followed by multiple ResNet blocks. 
% It produces the latent representation corresponding to the input image. 
The latent code $z$ is composed by two parts: the content part, $(z)_C$, which holds information about the image
content~(e.g. objects, position, scale, etc.), and the style part, $(z)_S$, which encodes the style that the content is presented in~(e.g. level of detail, shapes, etc.). 
The style component of the latent code $(z)_S$ is further split equally into 
$(z)_S = [(z)_S^{loc}, (z)_S^{glob}]$. 
Here, $(z)_S^{loc}$ encodes local style information per pixel of each feature map, while $(z)_S^{glob}$ undergoes further downsampling via a small sub-network to generate a single value per feature map.%\todo{perhaps we can emphasize better the global and local thing, need second pass for this}
\begin{figure*}[ht!]
\centering
\includegraphics[scale=0.375]{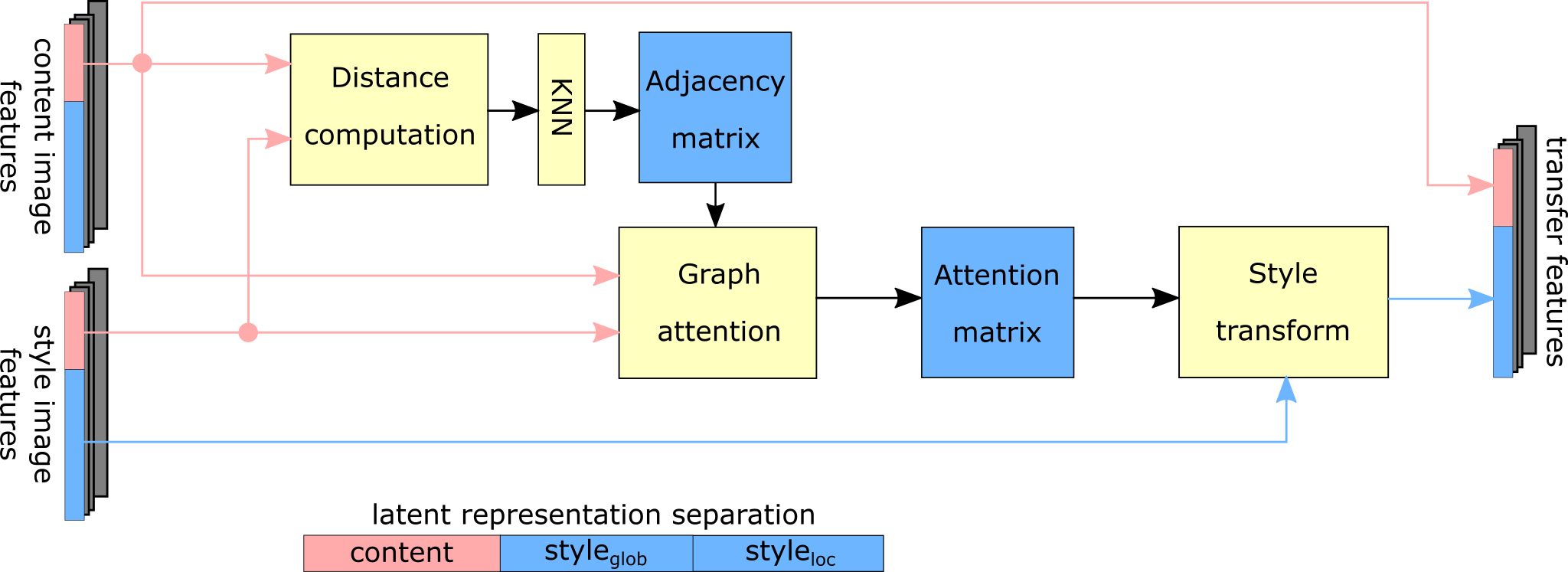}
\caption{One step of peer regularization takes as input latent representations of content and style images. The content part of the latent representation is used to induce a graph structure on the style latent space, which is then used to recombine the style part of the content image's latent representation from the style image's latent representation. This results in a new style latent code. The Two-Stage Peer Regularization Layer performs the peer regularization operation twice. In the second step, the roles of content and style information are swapped.}
\vspace{-2mm}
\label{fig:peer_layer}
\end{figure*}

\paragraph{Auxiliary decoder.} The Auxiliary decoder reconstructs an image from its latent representation and is used only during training to train the encoder module. 
It is composed of several ResNet blocks followed by fractionally-strided convolutional layers
to reconstruct the original image. 
The loss is inspired by~\cite{Sanakoyeu2018StyleAware,Kotovenko2019StyleDisentangle} and is composed of the following parts. A content feature cycle loss that pulls together latent codes representing the same content:
\begin{equation}
\begin{split} \label{eq:metric_loss_cont}
    L_{z_{cont}} &= f[E(D(T(z_i, z_t)))_C - (z_i)_C] \\
    &+ f[E(D(T(z_i, z_i)))_C - (z_i)_C]
\end{split}
\end{equation}

A metric learning loss, enforcing clustering of the style part of the latent representations:
\begin{equation}
\label{eq:metric_loss_style}
\begin{split}
    L_{z_{style}}^{pos} &= f[(z_{i_1})_S - (z_{i_2})_S] + f[(z_{t_1})_S - (z_{t_2})_S]  \\
    L_{z_{style}}^{neg} &= f[(z_{i_1})_S - (z_{t_1})_S] + f[(z_{i_2})_S - (z_{t_2})_S]  \\
    L_{z_{style}} &= L_{z_{style}}^{pos} + max(0.0, \mu - L_{z_{style}}^{neg}).
\end{split}
\end{equation}

A classical reconstruction loss used in autoencoders, which forces the model to learn perfect reconstructions of its inputs:
%\begin{split}
%    \tilde{L}_{z_{cycle}} &= f[E(\tilde{D}(z_i))_C - (z_i)_C] + f[E(\tilde{D}(z_i))_S - (z_i)_S] \\
%    &+ f[E(\tilde{D}(z_t))_S - (z_t)_S] \\
%    &+ f[E(\tilde{D}(z^{\prime}_i)) - z^{\prime}_i] + f[E(\tilde{D}(z^{\prime}_t)) - z^{\prime}_t] 
%\end{split}
\begin{equation}
\tilde{L}_{idt} = f[\tilde{D}(E(x_i)) - x_i] + f[\tilde{D}(E(x_t)) - x_t].
\end{equation}
And a latent cycle loss enforcing the latent codes of the inputs to be the same as latent codes of the reconstructed images:
\begin{equation}
\tilde{L}_{z_{cycle}} = f[E(\tilde{D}(z_i)) - (z_i)] + f[E(\tilde{D}(z_t)) - (z_t)],
\end{equation}
which has experimentally shown to stabilize the training.
The total auxiliary decoder loss $L_{\tilde{D}}$, is then defined as:
\begin{equation}
L_{\tilde{D}} = L_{z_{cont}} + L_{z_{style}} + \tilde{L}_{z_{cycle}} + \lambda \tilde{L}_{idt},
\end{equation}
where $\lambda$ is a hyperparameter weighting importance of the reconstruction loss $\tilde{L}_{idt}$ with respect to the rest.

\paragraph{Main decoder.} This network replicates the architecture of the auxiliary decoder, and uses 
the output of the Two-stage Peer-regularized Feature Recombination module~(see Section~\ref{sec:PeerRegFeatTransf}). 
During training of the main decoder the encoder is kept fixed, and the decoder is optimized using a loss function composed of the following parts. First, the decoder adversarial loss:
\begin{equation}
\begin{split}
L_{gen}&=\mathbb{E}_{x_{i} \sim \mathbb{P}}\left[\left(C\left(x_{i}\right)-\mathbb{E}_{x_{f} \sim \mathbb{Q}} C\left(x_{f}\right) + 1\right)^2 \right] \\
&+\mathbb{E}_{x_{f} \sim \mathbb{Q}}\left[\left(\mathbb{E}_{x_{i} \sim \mathbb{P}} C\left(x_{i}\right) - C\left(x_{f}\right) + 1\right)^2 \right],
\end{split}
\end{equation}
where $\mathbb{P}$ is the distribution of the real data and $\mathbb{Q}$ is the distribution of the generated~(fake) data and $C$ is the discriminator.

In order to enforce the stylization preserve the content part of the latent codes while recombining the style part of the latent codes to represent the target style class, we use so called transfer latent cycle loss:
\begin{equation}
\begin{split}
L_{z_{transf}} &= f[E(D(T(z_i, z_t)))_C - (z_i)_C] \\ 
&+ f[E(D(T(z_i, z_t)))_S - (z_t)_S].
\end{split}
\end{equation}
%\begin{split}
%    L_{z_{cycle}} &= f[E(D(z_i, z_t))_C - (z_i)_C] + f[E(D(z_i, z_i))_C - (z_i)_C] \\
%    &+ f[E(D(z_i, z_t))_S - (z_t)_S] + f[E(D(z_i, z_i))_S - (z_i)_S] \\ 
%    &+ f[E(D(z_t, z_t))_S - (z_t)_S]  \\
%    &+ f[E(D(z^{\prime}_i, z^{\prime}_i)) - z^{\prime}_i] + f[E(D(z^{\prime}_t, z^{\prime}_t)) - z^{\prime}_t] 
%\end{split}
Further, to make the main decoder learn to reconstruct the original inputs, we employ the classical reconstruction loss as we did for auxiliary decoder as well:
\begin{equation}
L_{idt} = f[D(T(z_i, z_i)) - x_i] + f[D(T(z_t, z_t)) - x_t].
\end{equation}
The above put together composes the main decoder loss $L_{D}$, where the reconstruction loss $L_{idt}$ is weighted by the same hyperparameter $\lambda$ used also in $L_{\tilde{D}}$.
\begin{equation}
L_{D} = L_{gen} + L_{z_{transf}} + \lambda L_{idt}
\end{equation}

\paragraph{Discriminator.} The discriminator is a convolutional network receiving two images concatenated over 
the channel dimension and producing an $N \times N$ map of predictions. 
The first image is the one to discriminate, whereas the second one serves as conditioning for the
style class. 
The output prediction is ideally $1$ if the two inputs come from the same style class and $0$ otherwise. The discriminator loss is defined as:
\begin{equation}
\begin{split}
L_{C}&=\mathbb{E}_{x_{i} \sim \mathbb{P}}\left[\left(C\left(x_{i}\right)-\mathbb{E}_{x_{f} \sim \mathbb{Q}} C\left(x_{f}\right) - 1\right)^2 \right] \\
&+\mathbb{E}_{x_{f} \sim \mathbb{Q}}\left[\left(\mathbb{E}_{x_{i} \sim \mathbb{P}} C\left(x_{i}\right) - C\left(x_{f}\right) -1\right)^2 \right].
\end{split}
\end{equation}
%\todo{what is Q above?}

\subsection{Two-stage Peer-regularized Feature Recombination (TPFR)}
\label{sec:PeerRegFeatTransf}
%Considering the growing popularity of geometric deep learning~\cite{Bronstein2017Geometric}, 
The TPFR module draws inspiration from PeerNets~\cite{Svoboda2019} 
and Graph Attention Layer (GAT)~\cite{Velickovic2018} to
perform style transfer in latent space, taking advantage of the separation of content and style information (enforced by Equations \ref{eq:metric_loss_cont} and \ref{eq:metric_loss_style}). Peer-regularized feature recombination is done in two stages, as explained in the following paragraphs.

\paragraph{Style recombination.} It receives $z_i = [(z_i)_C, (z_i)_S]$ and $z_t = [(z_t)_C, (z_t)_S]$ as an input and computes the k-Nearest-Neighbors (k-NN) between $(z_i)_C$ and $(z_t)_C$ 
using the Euclidean distance to induce the graph of peers. % with adjacency matrix given by the k-NN result.
% within the latent space of $z_r$ and $z_c$.

Attention coefficients over the graph nodes are computed and used to recombine the style portion of 
$(z_{out})_S$ as convex combination of its nearest neighbors representations. 
The content portion of the latent code remains instead unchanged, resulting in:
$z_{out} = [(z_i)_C, (z_{out})_S]$.

% \paragraph{Attention over feature graph.}
% Inspired by PeerNets~\cite{Svoboda2019}, which introduced the Peer Regularization Layer~(PR), a variant of Graph 
% Attention Layer~(GAT)~\cite{Velickovic2018}, our procedure aggregates style as follows. 
%\todo{this is not clear. where did we define xi? where is the style image defined?}
Given a pixel ${\mathbf{(z_m)_C}}$ of feature map $m$, its $k$-NN graph in the space of $d$-dimensional feature maps 
of all pixels of all peer feature maps $n_k$ is considered.
The new value of the style part $(z)_S$ for the pixel is expressed as:
\begin{equation}
\label{eq:PR_layer}
(\mathbf{\tilde{z}}^{m}_{p})_S = \sum_{k=1}^K \alpha_{m n_k p q_k} (\mathbf{z}^{n_k}_{q_k})_S,
\end{equation}
\begin{equation}
\alpha_{m n_k p q_k} = \frac{\mathrm{LReLU}(\exp(a((\mathbf{z}^m_{p})_C,(\mathbf{z}^{n_k}_{q_k})_C)))
}{
\sum_{k'=1}^K \mathrm{LReLU}(\exp(a((\mathbf{z}^m_{p})_C,(\mathbf{z}^{n_{k'}}_{q_{k'}})_C )))
}
\end{equation}
where $a(\cdot,\cdot)$ denotes a fully connected layer mapping from $2d$-dimensional input to scalar output, and $\alpha_{m n_k p q_k}$ are attention scores measuring the importance of the $q_k$th pixel of feature map $n$ to the output $p$th pixel $\tilde{\mathbf{x}}_p^m$ of feature map $m$. 
The resulting style component of the input feature map $\tilde{\mathbf{X}}^m$ is the weighted pixel-wise average of its peer pixel features defined over the style input image. 

\paragraph{Content recombination. } Once the style latent code is recombined, an analogous process is repeated to transform the content latent code according to the new style information. In this case, it starts off with inputs $z_{out} = [(z_i)_C, (z_{out})_S]$ and $z_t = [(z_t)_C, (z_t)_S]$ and the \textit{k}-NN graph is computed given the style latent codes $(z_{out})_S, (z_t)_S$. This graph is used together with Equation~\ref{eq:PR_layer} in order to compute the attention coefficients and recombine the content latent code as $(z_{final})_C$. 

The output of the TPFR module therefore is a new latent code $z_{final} = [(z_{final})_C, (z_{out})_S]$ which recombines both style and content part of the latent code.

% \subsection{Training}
% Our network can be trained end-to-end alternating optimization steps for the auxiliary generator, 
% the main generator, and the discriminator.

% The dataset of~\cite{Zhu2017CycleGAN}, composed of a collection of photographs and four different painter
% collections is used for training the model. 
% In particular, the datasets named \textit{monet2photo}, \textit{cezzane2photo}, \textit{vangogh2photo} and \textit{ukiyoe2photo} are combined into a single dataset named \textit{painter2photo}, consisting
% of 6280 real photos and 2560 paintings in total.

% The loss used for training is defined as:
% \begin{equation}
%     L = L_{D} + L_{G} + L_{\tilde{G}},
% \end{equation}
% where $D$ is the discriminator, $G$ the main generator, and $\tilde{G}$ is the auxiliary generator (see
% Section~\ref{subsec:Arch}).
% ADAM~\cite{Kingma2014Adam} is used as the optimizer with learning rate set to 0.0002 and batch size of 1, training is performed for total a of 200 epochs. 
% %\todo{this belongs more to experiments setup.}
% In each epoch, all the real photos are visited, which results in $~6280$ iterations per epoch. The weighting of the reconstruction identity loss $\lambda = 25.0$ and the margin for the metric learning $\mu = 4.0$ during all of our experiments. The training images are cropped and resized to $256 \times 256$ pixels. 
% Note that during testing, our method can operate on images of arbitrary size.

\begin{figure*}[ht!]
\centering
\includegraphics[scale=0.065]{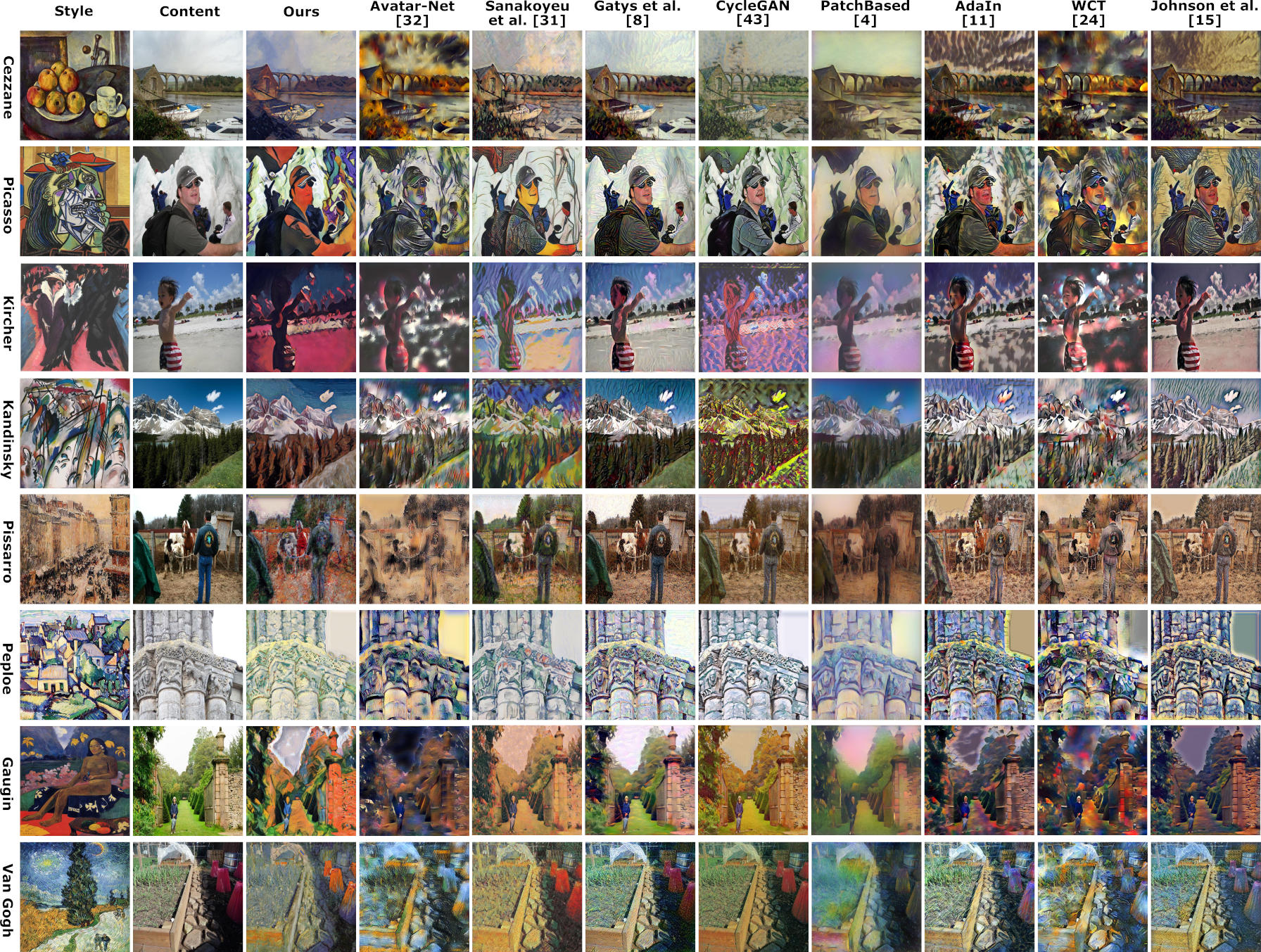}
\caption{Qualitative comparison with respect to other state-of-the-art methods. It should be noted that most of the compared methods had to train a new model for each style. While providing competitive results, our method performs arbitrary style transfer using a single model.}
%\todo{fill in text here, on what data etc. what can be noticed?}
\vspace{-1mm}
\label{fig:results_comparison}
\end{figure*}

\section{Experimental setup and Results}
The proposed approach is compared against the state-of-the-art on extensive qualitative evaluations and, to
support the choice of architecture and loss functions, ablation studies are performed to demonstrate roles of
the various components and how they influence the final result.
Only qualitative comparisons are provided as no standard quantitative evaluations are 
defined for NST algorithms.

\subsection{Training}
A modification of the dataset collected by~\cite{Sanakoyeu2018StyleAware} is used for training the model. It is composed of a collection of thirteen different painters representing different target styles and selection of relevant classes from the Places365 dataset~\cite{Zhou2017Places365} providing the real photo images.  
This gives us 624,077 real photos and 4,430 paintings in total.
Our network can be trained end-to-end alternating optimization steps for the auxiliary decoder, 
the main decoder, and the discriminator. The loss used for training is defined as:
\begin{equation}
    L = L_{C} + L_{D} + L_{\tilde{D}},
\end{equation}
where $C$ is the discriminator, $D$ the main decoder, and $\tilde{D}$ is the auxiliary decoder (see
Section~\ref{subsec:Arch}).
Using ADAM~\cite{Kingma2014Adam} as the optimization scheme, with an initial learning rate of $4\mathrm{e}{-4}$ and a batch size of 2, training is performed for a total of 200 epochs. After 50 epochs, the learning rate is decayed linearly to zero. Please note that choice of the batch size can be arbitrary, and we choose 2 only due to limited computing resources. 
%\todo{this belongs more to experiments setup.}
In each epoch, we randomly visit $6144$ of the real photos. The weighting of the reconstruction identity loss $\lambda = 25.0$ and the margin for the metric learning $\mu = 1.0$ during all of our experiments. The training images are cropped and resized to $256 \times 256$ resolution. 
Note that during testing, our method can operate on images of arbitrary size.
% To compare our approach to the state-of-the-art, extensive qualitative evaluation is performed. And to support the choice of architecture and loss functions, ablation studies showing the role of different components of our solution and their influence towards the final result are shown. Only qualitative comparison is provided, as there exist no common quantitative evaluations defined for NST algorithms.
%showing comparable results of our method with respect to the current state-of-the-art in NST. 

%\subsection{Style transfer}
\subsection{Style transfer} 
A set of test images from Sanakoyeu~\cite{Sanakoyeu2018StyleAware} is stylized and compared against competing 
methods in Figure~\ref{fig:results_comparison} (inputs of size $768 \times 768$ px) to demonstrate arbitrary
stylization of a content image given several different styles. It is important to note that, as opposed to majority of the competing methods, our network does not require retraining for each style and allows therefore also for transfer of previously unseen styles, e.g. Pissarro in row 5 of Figure~\ref{fig:results_comparison}, which is a painter that has not been present in the training set.

\begin{figure*}[ht!]
\centering
\subfigure[Zero-shot style transfer.]{
\includegraphics[scale=0.105]{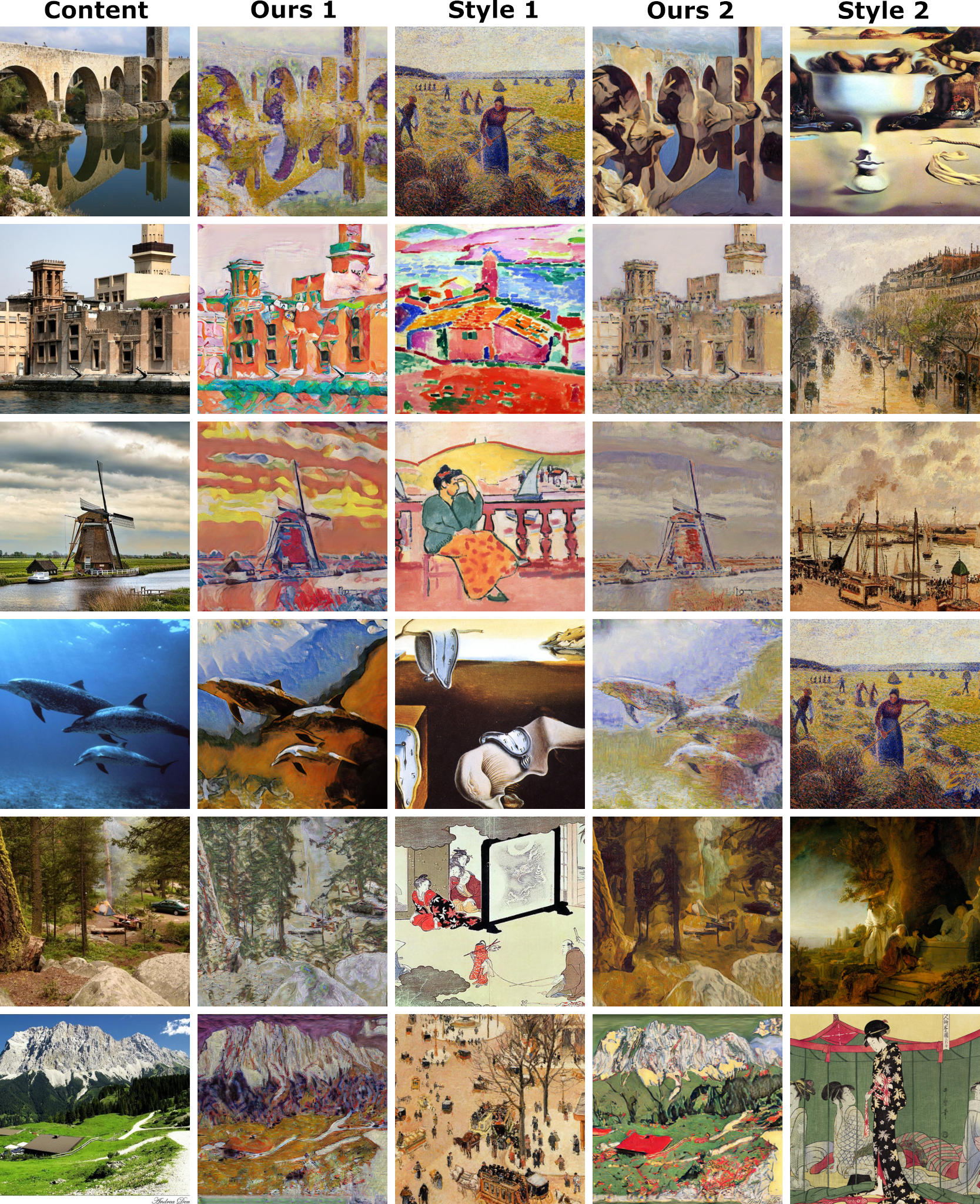}
\label{fig:results_zero_styles}
}
\hspace{1mm}
\subfigure[Styles seen during training.]{
\includegraphics[scale=0.105]{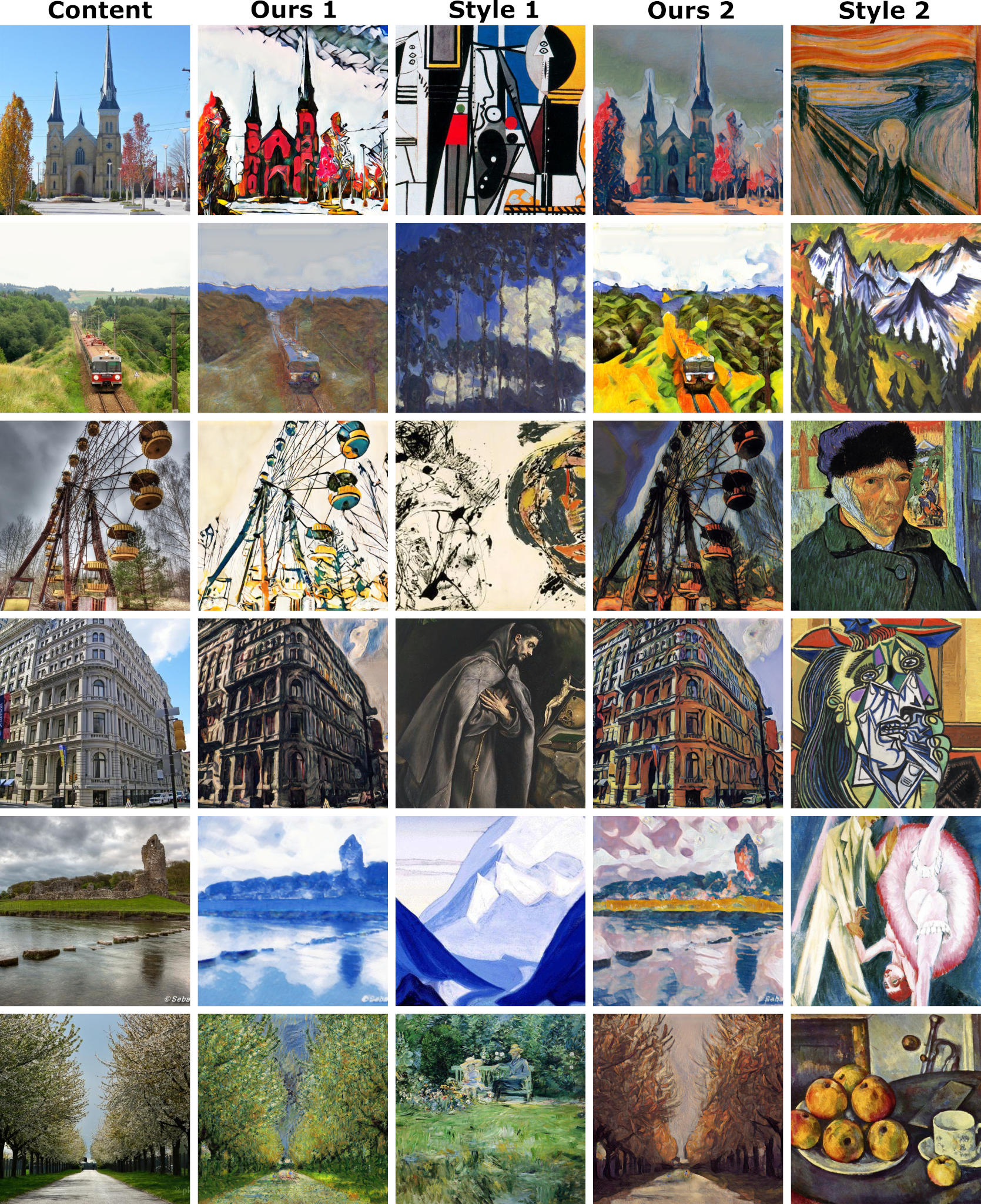}
\label{fig:results_arb_styles}
}
\caption{Qualitative evaluation of our method for previously unseen styles (left) and for styles seen during
training (right). It can be observed that the generated images are consistent with the provided target style
(inferred from a single sample only), showing the good generalization capabilities of the approach.}
%\vspace{-1mm}
\end{figure*}

Qualitative results\footnote{More results are shown in the supplementary material.} that were done on color images of size $512 \times 512$, using previously unseen paintings from painters that are in the training set, are shown in Figure~\ref{fig:results_arb_styles}. It is worth noticing that our approach can deal also with very abstract painting styles, such as the one of Jackson Pollock~(Figure~\ref{fig:results_arb_styles}, row 3, style 1). It motivates the claim that our model generalizes well to many different styles, allowing zero-shot style transfer. 

\paragraph{Zero-shot style transfer.} In order to support our claims regarding zero-shot style transfer, we have collected samples of a few painters that were not seen during training. In particular, we collected paintings from Salvador Dali, Camille Pissarro, Henri Matisse, Katigawa Utamaro and Rembrandt. The evaluation presented in Figure~\ref{fig:results_zero_styles} shows that our approach is able to account for fine details in painting style of Camille Pissarro~(row 1, style 1), as well as create larger flat regions to recombine style that is own to Katigawa Utamaro~(row 6, style 2). We find the results of zero-shot style transfer using the aforementioned painters very encouraging.

%Contrary to all the other methods, which typically use a VGG pre-trained model for the perceptual loss computation, our method allows for end-to-end training, which is of a major advantage due to the problems of image classification networks pointed out by Geirhos etal.~\cite{Geirhos2018ImageNetTrained}.

% \begin{figure}[ht!]
% \centering
% \includegraphics[scale=0.19]{images/random_styles_array.png}
% \caption{Qualitative evaluation of randomly coupled content and style images from the test set containing only painter styles seen during training.}
% \label{fig:random_styles}
% \end{figure}

\begin{figure*}[ht!]
\centering
\includegraphics[scale=0.115]{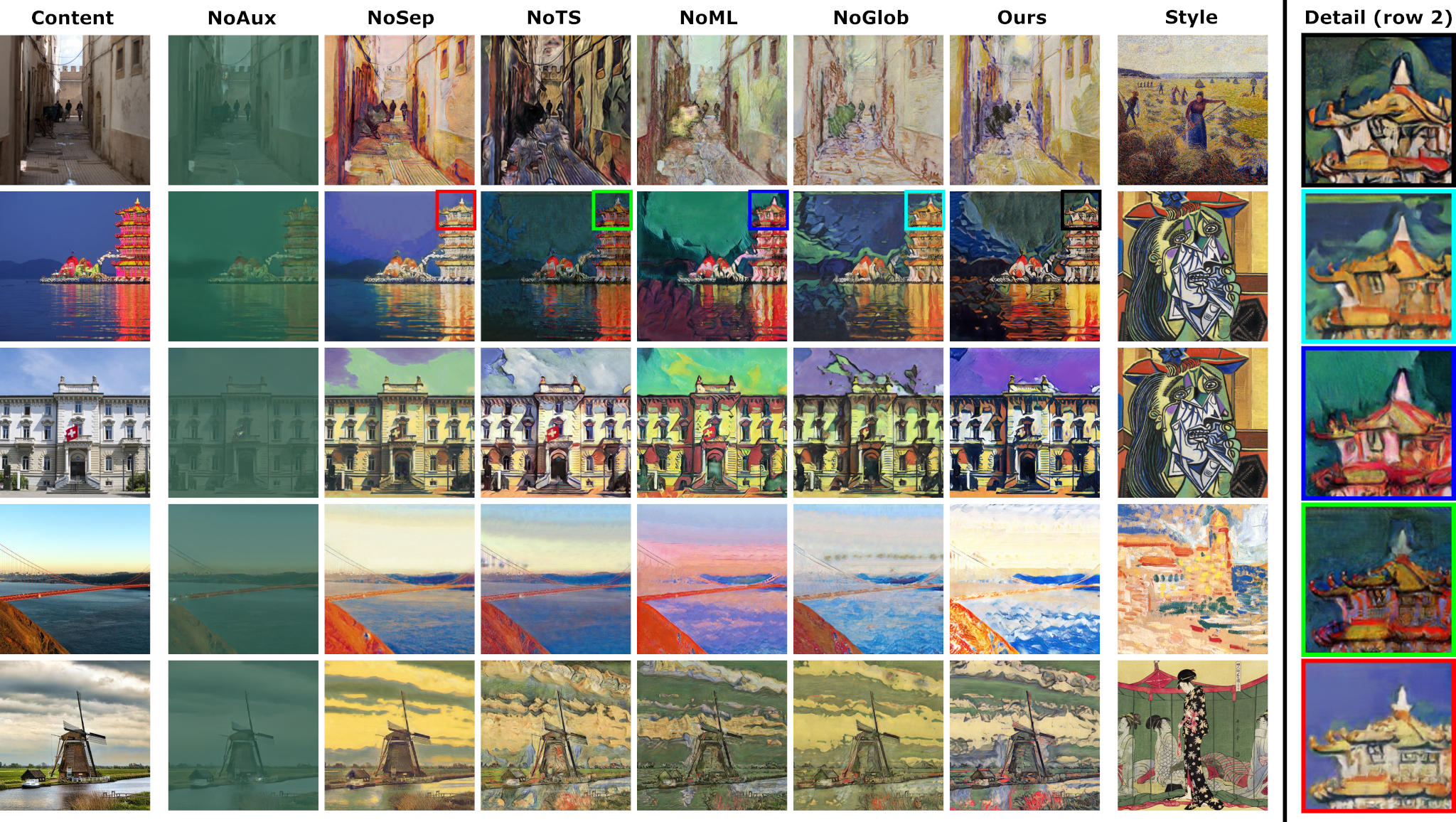}
\caption{Ablation study evaluating different architecture choices for our approach. Detail of style transfer for row 2 is shown in the rightmost column.
\textit{Ours} refers to the final approach, \textit{NoAux} makes no use of the auxiliary decoder during training, \textit{NoSep} ignores the separation of content and style in the latent space during feature exchange in TPFR module and exchanges the whole latent code at once, \textit{NoTS} recombines only style features based on content and keeps the original content features, \textit{NoML} does not use metric learning and \textit{NoGlob} does not separate the style latent code into a global and local part.
}

% The meaning of the rows is: \textit{Ours} - our final approach. \textit{NoSep} - ignoring the separation of content and style in the latent space during feature exchange in PeRFeaT layer. \textit{NoML} - no metric learning loss during training. \textit{NoAux} - no auxiliary decoder during training.}
\vspace{-5mm}
\label{fig:results_array_ablation}
\end{figure*}

%\subsection{Ablation study}
\paragraph{Ablation study.}
There are several key components in our solution which make arbitrary style transfer with a single model and end-to-end training possible. The effect of suppressing each of them during the training is examined, and results for the various models are compared, highlighting the importance of each component in Figure~\ref{fig:results_array_ablation}. Clearly, the \textit{auxiliary decoder} used during training is a centerpiece of the whole approach, as it prevents degenerate solutions. We observe that training encoder directly with the main decoder end-to-end does not work~(Fig.~\ref{fig:results_array_ablation} \textit{NoAux}). \textit{Separation of the latent code} into content and style part allows for the introduced two-stage style transfer and is important to account for changes in shape of the objects for styles like, e.g. Picasso~(Fig.~\ref{fig:results_array_ablation} \textit{NoSep}). \textit{Two-stage recombination} provides better generalization to variety of styles~(Fig.~\ref{fig:results_array_ablation} \textit{NoTS}). Performing only exchange of style based on content features completely fails in some cases~(e.g. row 1 in Fig.~\ref{fig:results_array_ablation}). Next, \textit{metric learning} on the style latent space enforces its better clustering and enhances some important details in the stylized image~(Fig.~\ref{fig:results_array_ablation} \textit{NoML}). Last but not least, the combined \textit{local and global} style latent code is important in order to be able to account for changes in edges and brushstrokes appropriately~(Fig.~\ref{fig:results_array_ablation} \textit{NoGlob}). 

%\paragraph{Separation of latent space.} The second row of Figure~\ref{fig:results_array_ablation} shows that exchanging the entire latent code (content + style) will destroy much of the content information. Without the separation of the latent space into content and style, important features defining the content of the image will not be preserved and the resulting image will hardly resemble the originally desired content.

%\paragraph{Metric learning.} Arbitrary style transfer is possible due to metric learning on the style latent space. The results of omitting a metric-learning loss are shown in third row of Figure~\ref{fig:results_array_ablation}. One can clearly see the model is unable to make any use of the input style information and a very constant stylization is being generated.

%\paragraph{Auxiliary decoder.} The auxiliary decoder ensures that the latent code does not go ignored during reconstruction. If we train the GAN-based encoder-decoder model directly without an auxiliary decoder training the encoder, much of the stylization is learned by the decoder~(see last row of Figure~\ref{fig:results_array_ablation}). It mostly ignores the latent information, no matter the style image, again resulting in very constant stylization.

\section{Conclusions}
We propose a novel model for neural style transfer which mitigates various limitations of current 
state-of-the-art methods and that can be used also in the challenging zero-shot transfer setting.
This is thanks to a Two-Stage Peer-Regularization Layer using graph convolutions to recombine the style
component of the latent representation and with a metric learning loss enforcing separation of
different styles combined with cycle consistency in feature space.
An auxiliary decoder is also introduced to prevent degenerate solutions and to enforce 
enough variability of the generated samples.
The result is a state-of-the-art method that can be trained end-to-end without the need of a pre-trained
model to compute the perceptual loss, therefore lifting recent concerns regarding the reliability of such features
for NST. 
More importantly the proposed method requires only a single encoder and a single decoder to perform
transfer among arbitrary styles, contrary to many competing methods requiring a decoder (and possibly an encoder) 
for each input and target pair. This makes our method more applicable to real-world image generation scenarios
where users define their own styles.
%\todo{check conclusions}
% 
% contrary to other methods where multiple networks are required.

% This paper proposes a novel method for arbitrary neural style transfer, which mitigates various limitations of state-of-the-art methods. Taking advantage of the Peer-Regularization approach, we propose an adversarial model that performs the style transfer in latent space, separating style and content information, such that image content is preserved while style is altered. An auxiliary decoder strategy is applied to avoid posterior collapse, and the Relativistic GAN formulation is used to stabilize the GAN training and achieve visually-pleasing results. Contrary to other works, our model does not need the widely-used perceptual loss. This avoids any further concerns in the NST community regarding the reliability of features from image classification models. Our model can be trained end-to-end, and the results are competitive among the current state-of-the-art. Moreover, it shows satisfying performance on previously unseen styles, resulting in zero-shot style transfer.

%\subsubsection*{Acknowledgments}
%Use unnumbered third level headings for the acknowledgments. All acknowledgments go at the end of the paper. Do not include acknowledgments in the anonymized submission, only in the final paper.

{\small
\bibliographystyle{ieee_fullname}
\bibliography{egbib}
}

\clearpage

\appendix

\section{Network architecture}
This section describes our model in detail. We describe the encoder-decoder network and discriminator in separate sections below. We provide our code containing the implementation details~\footnote{Code is available at this link: http://nnaisense.com/conditional-style-transfer} to assure full reproducibility of all the presented results.
\subsection{Autoencoder}
\begin{figure*}[ht!]
\centering
\includegraphics[scale=0.325]{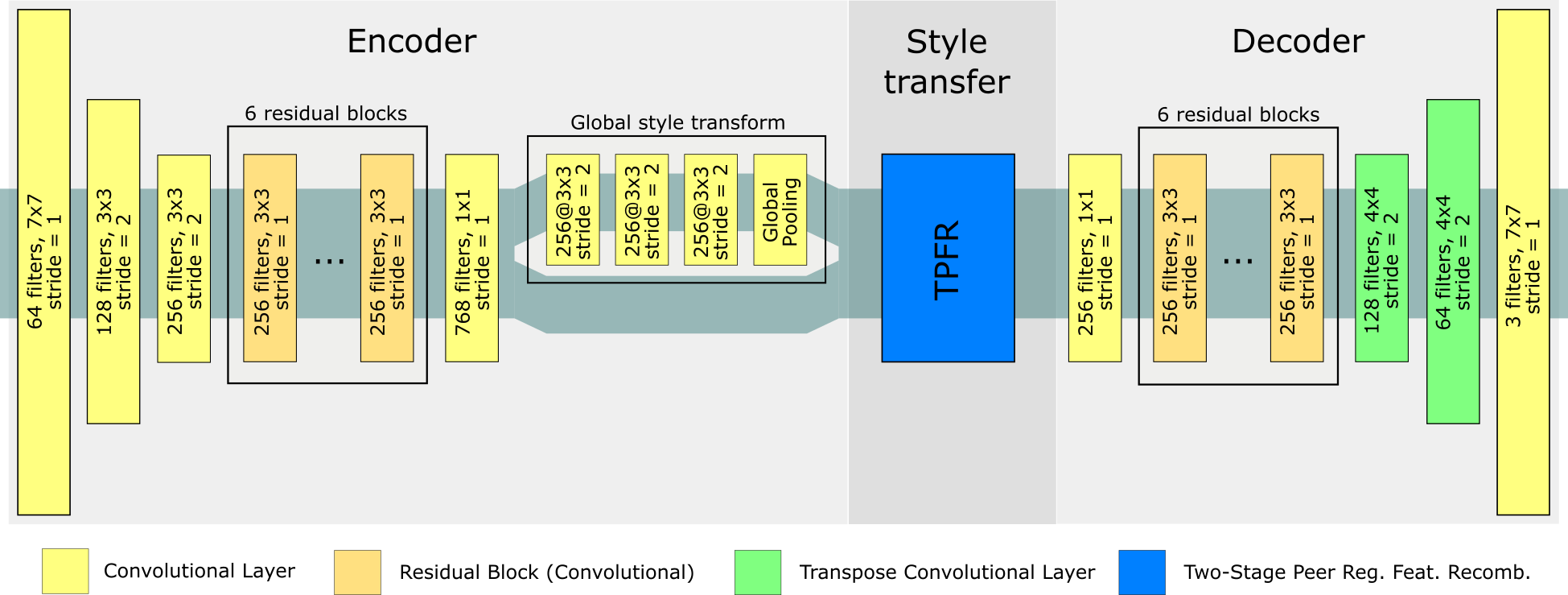}
\caption{Detailed architecture of autoencoder. The faded blue line in the background shows the flow of information through the model.}
\label{fig:arch_gen}
\end{figure*}

Detailed scheme of the architecture is depicted in Figure~\ref{fig:arch_gen}. Each of the convolutional layers~(in yellow) is followed by Instance Normalization~(IN)~\cite{Ulyanov17IN} and ReLU nonlinearity~\cite{Nair2010ReLU}. The TPFR module uses a variant of the Peer Regularization Layer~\cite{Svoboda2019} with Euclidean distance metric, k-NN with $K=5$ nearest neighbours and dropout on the attention weights of $0.2$.

The generated latent code are $768$ feature maps of size $(W / 4) \times (H / 4)$, where $W$ and $H$ are the input width and height respectively. First $256$ feature maps is the content latent representation, while the remaining $512$ is for the style. The style latent representation is further split into halves, having first $256$ feature maps left unchanged and the second $256$ feature maps are passed through the \textit{Global style transform} block producing feature maps of size $1 \times 1$ that hold the global part of the style latent representation.

The last convolutional block of the decoder is equipped with TanH nonlinearity and produces the reconstructed RGB image.

The auxiliary decoder copies the architecture of the main decoder, while omitting the \textit{Style transfer} block (see Figure~\ref{fig:arch_gen}).

\subsection{Discriminator}
\begin{figure}[ht!]
\centering
\includegraphics[scale=0.325]{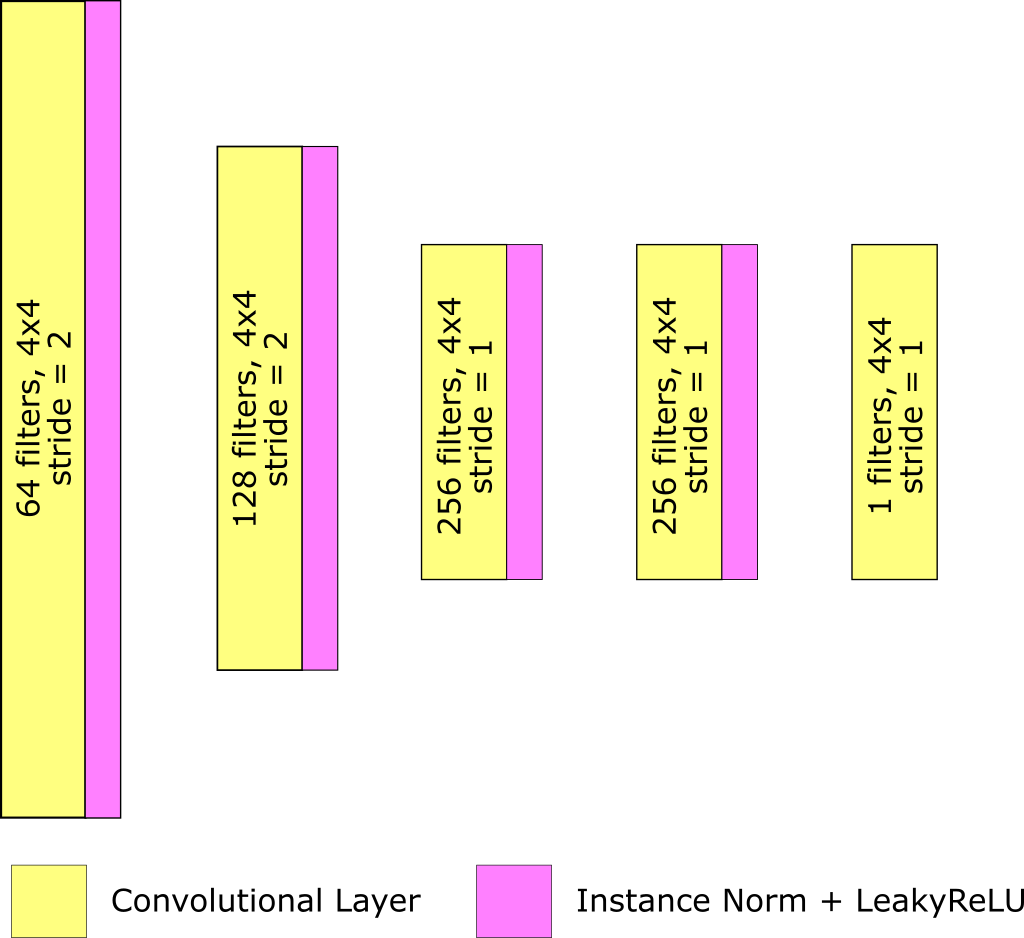}
\caption{Detailed architecture of discriminator.}
\label{fig:arch_disc}
\end{figure}

The discriminator architecture is shown in Figure~\ref{fig:arch_disc}. It takes two RGB images concatenated over the channel dimension as input and produces a $(W / 4) \times (H / 4)$ map of predictions. Our implementation uses LS-GAN and therefore there is no Sigmoid activation at the output

To stabilize the discriminator training, we add random Gaussian noise to each input:
\begin{equation}
    X = X + N(\mu, \sigma),
\end{equation}
where $N$ is a Gaussian distribution with mean $\mu = 0$ and standard deviation $\sigma = 0.1$.

\section{Style transfer results}
\begin{figure*}[ht!]
\centering
\includegraphics[scale=0.11]{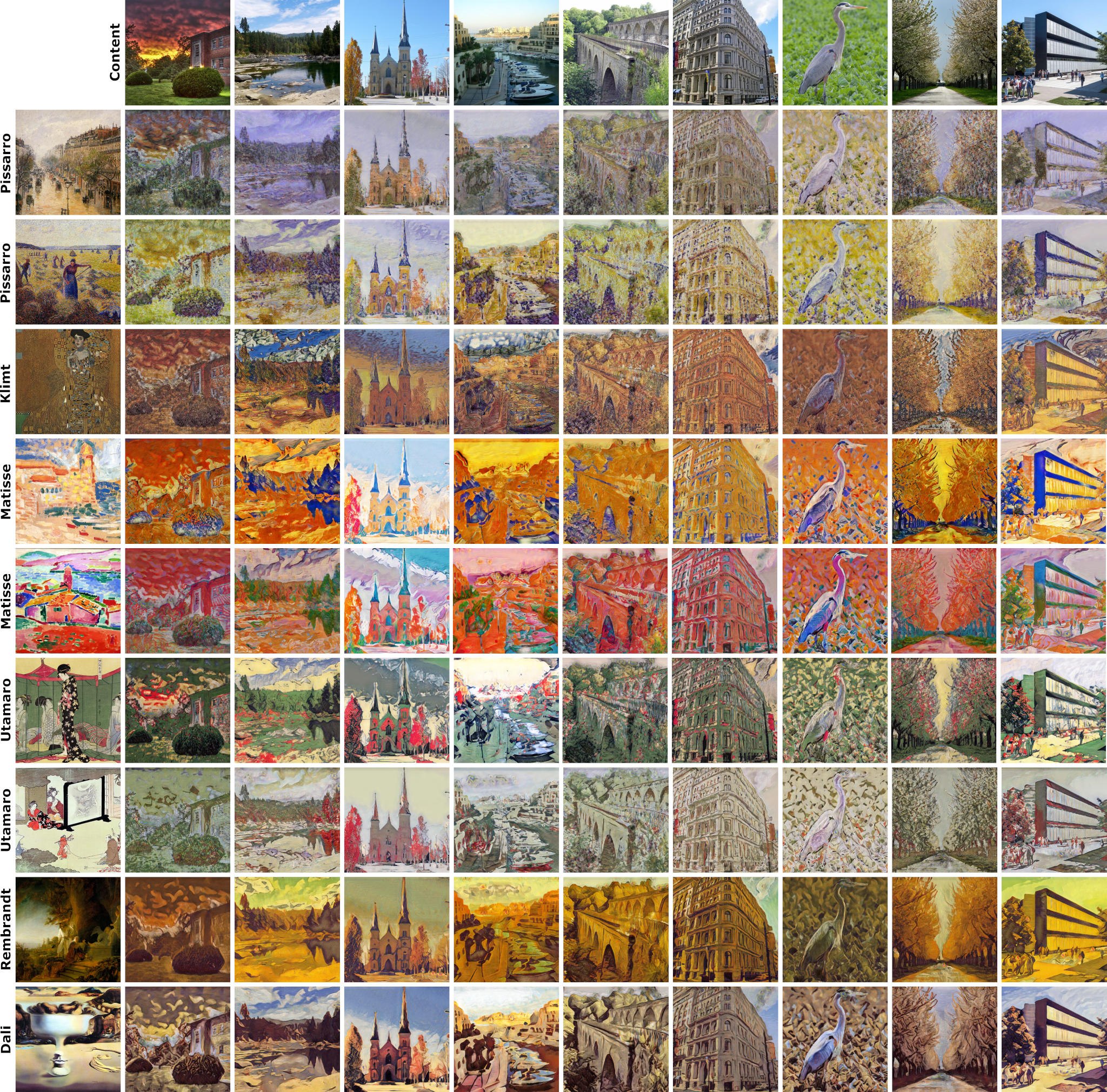}
\caption{Qualitative evaluation of our method in zero-shot style transfer setting. The results clearly show that our method generalizes well to previously unseen styles.}
\label{fig:zero_styles}
\end{figure*}

This section provides more qualitative results of our style transfer approach that did not fit in the main text. Figure~\ref{fig:zero_styles} are images generated with resolution $512 \times 512$ and shows the generalization of our approach to different styles and ability of our approach to perform zero-shot style transfer. In particular, we have collected some paintings from Salvador Dali, Camille Pissarro, Henri Matisse, Katigawa Utamaro and Rembrandt. 

\begin{figure*}[ht!]
\centering
\includegraphics[scale=0.11]{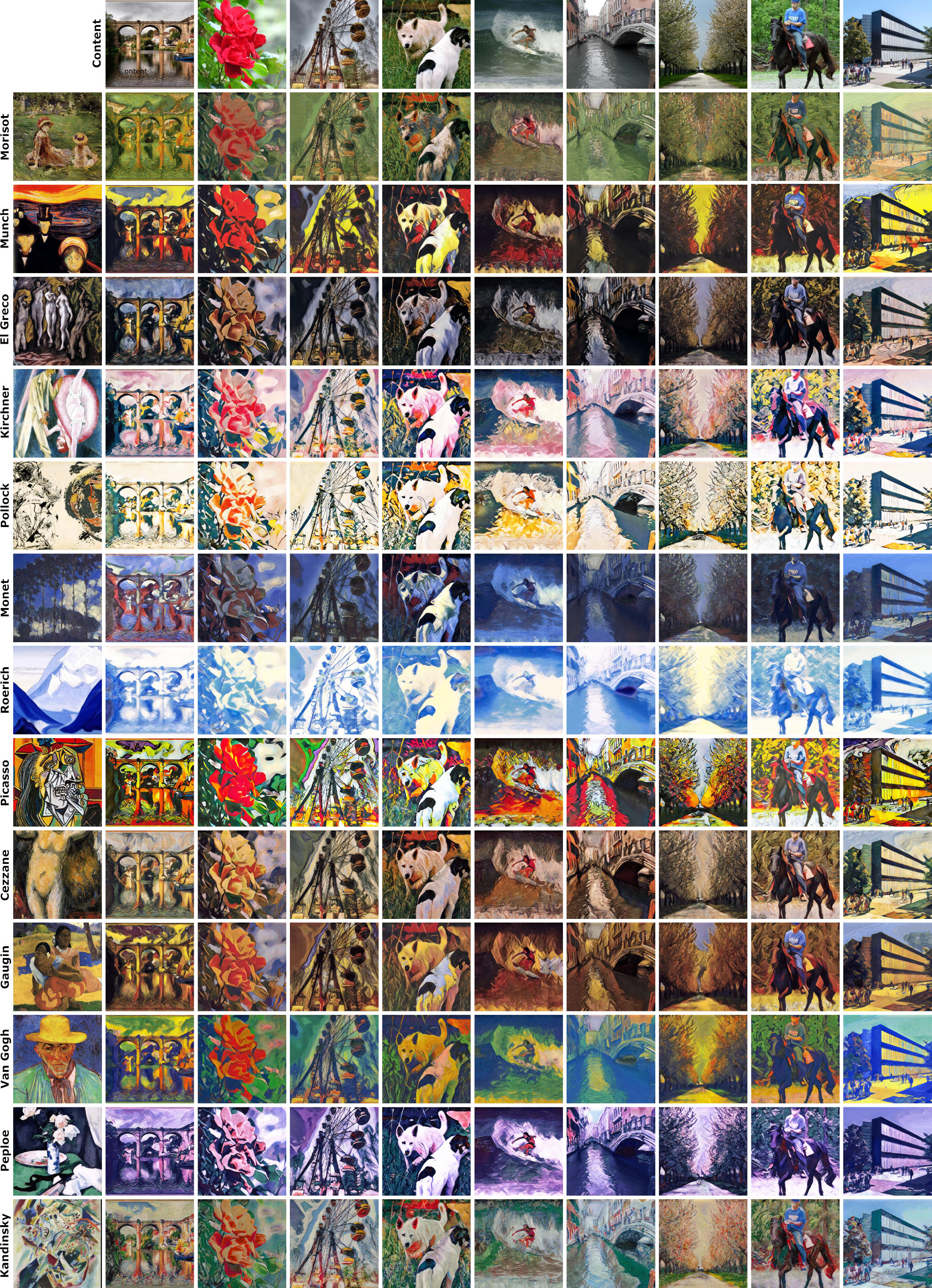}
\caption{Qualitative evaluation of our method using disjoint set of painting styles that were in the training set. }
\label{fig:arb_styles}
\end{figure*}

In addition, images in Figure~\ref{fig:arb_styles} were generated with resolution $256 \times 256$ and show results of transfer taking a random painting from the dosjoint test set of thirteen painting styles that our model was trained with~(Morisot, Munch, El Greco, Kirchner, Pollock, Monet, Roerich, Picasso, Cezzane, Gaugin, Van Gogh, Peploe and Kandinsky).

\section{Latent space structure} 
Our latent representation is split into two parts, $(z)_C$ and $(z)_S$, content and style respectively. Metric learning loss is used on the style part in order to enforce a separation of different modalities in the style latent space. 

\begin{equation}
\label{eq:metric_loss_style}
\begin{split}
    L_{z_{style}}^{pos} &= f[(z_{i_1})_S - (z_{i_2})_S] + f[(z_{t_1})_S - (z_{t_2})_S]  \\
    L_{z_{style}}^{neg} &= f[(z_{i_1})_S - (z_{t_1})_S] + f[(z_{i_2})_S - (z_{t_2})_S]  \\
    L_{z_{style}} &= L_{z_{style}}^{pos} + max(0.0, \mu - L_{z_{style}}^{neg}).
\end{split}
\end{equation}

where $(z_{i_1})_S$, $(z_{i_2})_S$ are style parts of latent representations of two different input images and $(z_{t_1})_S$, $(z_{t_2})_S$ are style parts of latent representations of two different targets from the same target class. Parameter $\mu = 1$ and it is the margin we are enforcing on the separation of the positive and negative scores.

\subsection{Visualization in image space}
\begin{figure*}[ht!]
\centering
\includegraphics[scale=0.115]{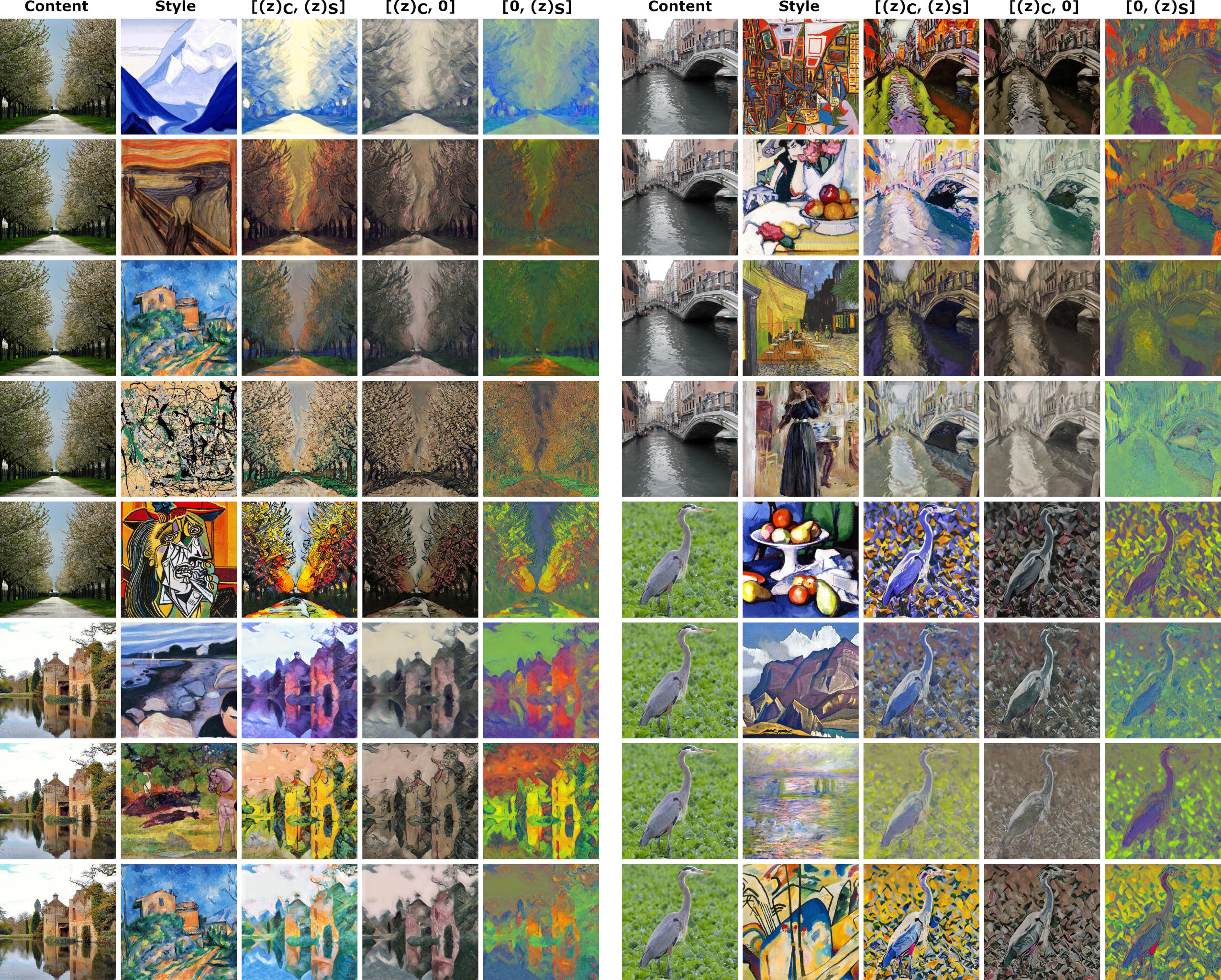}
\caption{Visualization of information contained in content and style parts of the latent representation. Even if $(z)_C$ is set to $0$, there is still some rough resemblance of the structure of the \textit{Content} image, because the TPFR module transforms a partially local style features based on the content features.}
\label{fig:ablation_reconst}
\end{figure*}

Figure~\ref{fig:ablation_reconst} visualizes the influence of the $(z)_C$ and $(z)_S$ parts of the latent representation after decoding back into the RGB image space. The TPFR module, which performs the style transfer, is executed first. The resulting latent code is then modified before feeding it to the decoder. Replacing the $(z)_C$ with $0$ gives us some rough representation of the style with only approximate shapes. On the other hand, if we replace $(z)_S$ with $0$ and we keep $(z)_C$, a rather flat representation of the input with sharper is reconstructed. This demonstrates that $(z)_C$ represents the content, while $(z)_S$ holds most of the style-related information. 

The fact that the latent code is passed through the TPFR module first means that the two-stage feature recombination is performed on the data we visualize. As a result,  the decoded image $[0, (z)_S]$ slightly resembles the structure of the content image even if the $(z)_C$ is set to $0$. Likewise, in case of $[(z)_C, 0]$, the geometry of the objects is already slightly modified based on the resulting style.

\section{Computational overhead}
Stylization of a single image of resolution $512\times512$ using our method takes approximately $16$ms on a single 
\textit{Titan-V100} GPU. Execution of the TPFR block takes approximately $3$ms, which is $18.75\%$ of the whole runtime. Due to memory requirements, our method can currently process images of size up to $768\times768$ pixels.

\section{Quantitative evaluation}
We are aware of the recent efforts bringing in quantities such as deception score~\cite{Sanakoyeu2018StyleAware} or content and style distribution divergence~\cite{Kotovenko2019StyleDisentangle}. However we decided not to use these metrics as they are all based on a VGG network trained to classify paintings. We argue that such evaluation may favor models that have used VGG perceptual losses for training. This concern is closely related to the work of~\cite{Geirhos2018ImageNetTrained}. 

\end{document}